\documentclass[runningheads]{llncs}
\usepackage{graphicx}
\usepackage[hidelinks]{hyperref}
\usepackage{cite}
\usepackage{tikz}
\usepackage{amsmath,bm}
\usepackage{dsfont}
\usepackage{siunitx}
\usepackage{algorithm2e}
\PassOptionsToPackage{hyphens}{url}\usepackage{hyperref}
\usepackage{amssymb}
\usepackage{todonotes}
\usepackage{float}
\usepackage{listings}
\usepackage[T1]{fontenc}
\definecolor{myblack}{RGB}{0, 0, 0}
\definecolor{mygreen}{RGB}{0, 146, 146}
\definecolor{myblue}{RGB}{0, 110, 219}
\definecolor{myred}{RGB}{146, 0, 0}

\newcommand{\CPP}{C\nolinebreak\hspace{-.05em}\raisebox{.4ex}{\tiny\bf +}\nolinebreak\hspace{-.10em}\raisebox{.4ex}{\tiny\bf + }}

\begin{document}
\title{GPRat: Gaussian Process Regression with Asynchronous Tasks}
\titlerunning{GPRat}
\author{Maksim Helmann \and
Alexander Strack\orcidID{0000-0002-9939-9044}
\and Dirk Pflüger \orcidID{0000-0002-4360-0212}}
\authorrunning{M. Helmann \and et. al}
%
\institute{Institute of Parallel and Distributed Systems, University of Stuttgart,\\ 70569 Stuttgart, Germany\\ 
\email{maksim.helmann@mail.utoronto.ca}\newline
\email{\{alexander.strack, dirk.pflueger\}@ipvs.uni-stuttgart.de}
}

\maketitle              
\begin{abstract}
Python is the de-facto language for software development in artificial intelligence (AI). 
Commonly used libraries, such as PyTorch and TensorFlow, rely on parallelization built into their BLAS backends to achieve speedup on CPUs. 
However, only applying parallelization in a low-level backend can lead to performance and scaling degradation.
In this work, we present a novel way of binding task-based \CPP code built on the asynchronous runtime model HPX to a high-level Python API using pybind11. 
We develop a parallel Gaussian process (GP) library as an application. 
The resulting Python library GPRat combines the ease of use of commonly available GP libraries with the performance and scalability of asynchronous runtime systems. 
We evaluate the performance on a mass-spring-damper system, a standard benchmark from control theory, for varying numbers of regressors (features). The results show almost no binding overhead when binding the asynchronous HPX code using pybind11. 
Compared to GPyTorch and GPflow, GPRat shows superior scaling on up to 64 cores on an AMD EPYC $7742$ CPU for training. 
Furthermore, our library achieves a prediction speedup of $7.63$ over GPyTorch and $25.25$ over GPflow.
If we increase the number of features from eight to 128, we observe speedups of $29.62$ and $21.19$, respectively.
These results showcase the potential of using asynchronous tasks within Python-based AI applications. 
\end{abstract}
\setcounter{footnote}{0} 
\section{Introduction}\label{sec:introduction}

Gaussian processes (GPs), also known as kriging, are a powerful regression technique widely used for modeling nonlinear relationships in various fields. Popular applications include geostatistics, control theory \cite{Kocijan2015}, and artificial intelligence (AI)\cite{Williams1995, Rasmussen2006}. 
GPs provide a fully probabilistic framework, allowing for uncertainty estimation through posterior variances, making them valuable in applications where the confidence of predictions is crucial \cite{Deisenroth2015, DoshiVelez2017, Wang2018}.
Popular Python GP libraries such as GPflow \cite{Matthews2017} and GPyTorch \cite{Gardner2018} leverage numerical libraries such as TensorFlow \cite{Abadi2015} and PyTorch \cite{Paszke2019} to achieve satisfactory performance and portability. 
While problem sizes are ever-increasing in the era of Big Data, development often focuses on additional features.
Performance-critical optimizations, such as parallelization and portability, are typically outsourced to the backends. 
If they do not integrate recent advances in high-performance computing (HPC), this can result in suboptimal resource usage.
A core advancement in parallel computing that is typically neglected is the advent of asynchronous runtime systems: They are more difficult to integrate than classical synchronous approaches. 
To address this challenge, this work introduces GPRat, a novel parallel library written in \CPP that leverages the HPX runtime system \cite{Kaiser2020} for efficient task-based parallelization.
By combining the performance of \CPP with an easy-to-use Python interface, GPRat ensures effortless integration into existing workflows.
The library takes advantage of HPX features like tasks futurization and dynamic task-graph scheduling with work-stealing \cite{Blumofe1999}.
Furthermore, GPRat utilizes a BLAS backend \cite{Blackford2002} to handle numerically intensive operations, further improving its performance.

Our main contributions in this work include:
\begin{itemize}
	\item  A novel, fully asynchronous task-based GP library with Python bindings that uses HPX, BLAS and multiple different tiled algorithms,
	\item A thorough performance comparison between GPRat and two reference libraries GPflow and GPyTorch in a node-level strong scaling test on an AMD EPYC $7742$ CPU, and
    \item A performance benchmark for varying problem sizes between GPRat and the reference implementations for $64$ cores on the same system.
\end{itemize}

The remainder of this work is organized as follows. 
Firstly, Section $2$ reviews related software frameworks for GPs, HPC, and Python bindings. 
Secondly, Section $3$ outlines the software components used in GPRat.
Subsequently, Section $4$ talks about the basic concepts of GPs, tiled algorithms, and the exposure of HPX to Python within GPRat.
Section $5$ presents the results of the comparison between GPRat and the two reference implementations.
Finally, Section $6$ concludes and outlines directions for future research and development.

\section{Related work}\label{sec:related_work}

In recent years, GPs have become increasingly popular in 
statistical modeling, driven by accessible and portable Python libraries like GPflow \cite{Matthews2017} and GPyTorch \cite{Gardner2018}. 
GPflow and GPyTorch excel due to their integration of TensorFlow \cite{Abadi2015} and PyTorch \cite{Paszke2019}, respectively. 
Both libraries leverage automatic differentiation and utilize BLAS backends for accelerated performance.
In this work, we use the Intel oneAPI Math Kernel Library (MKL) \cite{mkl} in GPflow, GPyTorch, and GPRat. Note that in GPflow, MKL is not activated per default. 
Enhancing computational performance often involves the use of message passing execution models, specifically the MPI standard \cite{Gropp1999}, implemented in frameworks such as OpenMPI \cite{Edgar2004}.
However, MPI-based approaches have the disadvantage of relying on, sometimes global, synchronization barriers. These barriers can hinder downstream computations that might otherwise continue asynchronously using partial results \cite{Badia2015, Grama2003}.
In shared memory environments, the OpenMP multiprocessing interface\cite{Dagum1998} is a popular alternative to MPI to remove obsolete communication overhead and simplify parallelization.
However, the fork-join computation model has similar synchronization requirements after parallel loops.
An alternative to MPI-based and fork-join approaches for HPC are task-based parallel programming models. 
In \cite{Thoman2018}, the authors provide a comparative analysis of existing task-based runtimes.
Among these, HPX \cite{Kaiser2020} stands out by adhering to the \CPP standard library API and additionally extending the basic functionality with support for accelerators and distributed computing (see Section \ref{sec:framework}).

A recent study \cite{Strack2023} demonstrates how HPX can accelerate GP regression by implementing a fully asynchronous tiled Cholesky decomposition to compute the inversion of the covariance matrix in Eq. \ref{eq:pred}. 
Their implementation uses sequential BLAS operations on a tiled matrix.
Their node-level benchmark, inter alia, demonstrates that HPX exhibits superior parallel scaling compared to the MPI-based approach in PETSc \cite{Balay2024}.
Based on their work, our work extends the framework by introducing additional functionality and Python bindings to create the GPRat library.

Although task-based parallelism can significantly improve performance, modern software developers are also increasingly seeking ``multi-language support'' to improve productivity \cite{Liegeois2023}. 
Python \cite{Perez2011}, a dynamic programming language, meets this need. 
It enables the integration of performance-critical code written in lower-level languages, such as \CPP. 
This is achieved through extension modules like pybind11 \cite{Wenzel2017}, which allow developers to expose \CPP functionality to Python without additional complexity.
One such approach is the PyKokkos \cite{Awar2022} library that enables performance-portable parallel computing in Python by automatically generating bindings between Python and the Kokkos C++ library using pybind11.
Similarly, efforts to combine Python with asynchronous tasks include the Phylanx platform \cite{Tohid2018}.
Phylanx allows to write Python or NumPy code and execute it with HPX. 
Therefore, the Python code is first transformed into an internal Phylanx representation and then into HPX tasks.
In contrast, we directly develop the code in \CPP and expose the desired functionality to Python via pybind11. 
Furthermore, we allow to manually manage the HPX runtime within Python.
This approach minimizes binding overhead and allows to maintain the performance of HPX.

\section{Software framework}\label{sec:framework}
The two main software tools that we use are the \CPP standard library for parallelism and concurrency HPX~\cite{Kaiser2020} and the pybind11 binding library \cite{Wenzel2017}.

\subsection{HPX}

HPX \cite{Kaiser2020}, developed by the STE||AR Group and first released in 2008, is a \CPP standard library designed for concurrency and parallelism.
Its API adheres to the \CPP11/14/17/20 ISO standards and complies with the programming principles of the Boost \CPP libraries. 
A standout feature of HPX is its ability to execute code fully asynchronous through HPX futures, avoiding induced synchronization barriers common in message-passing or fork-join programming models.
HPX's dataflow and future-based synchronization mechanisms make it easy to manage concurrency and build complex parallel tasks.
By leveraging an active global address space, HPX supports distributed computing \cite{Heller2019}.
HPX ensures portability across x86, ARM \cite{Diehl2023C}, and RISC-V \cite{Diehl_2023} architectures, as well as across accelerators from NVIDIA, AMD, and Intel \cite{Daiss2023}. Furthermore, it incorporates various static or dynamic schedulers that support work-stealing \cite{Blumofe1999}.

\subsection{pybind11}

pybind11 \cite{Wenzel2017} is a lightweight library that allows seamless integration between \CPP and Python. Developers can expose \CPP code to Python with minimal binding overhead, making it suitable for high-performance applications.
It supports modern \CPP standards (\CPP$11/14/17/20$) and provides automatic type conversion between the two languages, improving maintainability and reducing the need for manual conversions.
Furthermore, pybind11 integrates easily with build systems such as CMake and Setuptools, simplifying its incorporation into existing \CPP and Python projects.

\section{Methods}\label{sec:methods}

The next subsection first covers GP fundamentals. 
Then, we introduce the parallelization concept of tiled algorithms and BLAS tasks.
Lastly, we present the functionality of GPRat and describe the mechanism for initializing and terminating the HPX runtime.

\subsection{Gaussian processes}

GPs are a powerful framework for modeling complex, nonlinear phenomena, such as regression in supervised learning \cite{Gibbs1998}.
The objective is to determine a nonlinear function $f$ that maps input data from a feature matrix $Z$$=$$[\mathbf{z}_1, \mathbf{z}_2, \ldots, \mathbf{z}_n]^\top$, where $\mathbf{z}_i\in \mathbb{R}^D$, to observations $\mathbf{y}$$=$$[y_1, y_2, \ldots, y_n]^\top$, where $y_i \in \mathbb{R}$.
A GP extends the Gaussian distribution to infinite dimensions, where any finite set of random variables follows a joint Gaussian distribution \cite{Rasmussen2006}. 
It is completely defined by a mean function $m(\mathbf{z})$ and a covariance function, or kernel, $k(\mathbf{z}, \mathbf{z}')$
\begin{equation}
f(\mathbf{z}) \sim \mathcal{GP}\left(m(\mathbf{z}), k(\mathbf{z}, \mathbf{z}') \right).
\end{equation}
Typically, the mean function is assumed to be zero for simplicity \cite{Kocijan2015}.

In our work, we choose the squared exponential covariance function
\begin{equation}\label{eq:rbf}
	k\left(\mathbf{z}_i, \mathbf{z}_j \right) = \nu \cdot \exp \left[ -\frac{1}{2 \cdot l^2} \sum^D_{d=1} \left(z_{i,d} - z_{j,d} \right)^2 \right] + \delta_{ij} \sigma^2,
\end{equation}
where $D$ is the length of the feature vectors. The hyperparameters length scale $l$, vertical scale $\nu$, and noise variance $\sigma^2$ are crucial for the quality of the GP model. $\delta_{ij}$ denotes the Kronecker delta, which is $1$ if $i$$=$$j$ and $0$ otherwise.  
Using this kernel function, the covariance matrix $K \in \mathbb{R}^{N \times N}$ is constructed by evaluating all pairwise combinations of feature vectors.
The mean of the conditional distribution of $f\left(\hat{Z}\right)$, given the observations $\mathbf{y}$, provides the GP's predictions for the test inputs
\begin{equation}\label{eq:pred}
\hat{\mathbf{y}} = K_{\hat{Z}, Z} K^{-1}\mathbf{y},
\end{equation}
where $K_{\hat{Z}, Z} \in \mathbb{R}^{M \times N}$ is the cross-covariance matrix between the observed data $Z \in \mathbb{R}^{N \times D}$ and the test data $\hat{Z} \in \mathbb{R}^{M \times D}$. 
The prior covariance matrix $K_{\hat{Z}, \hat{Z}} \in \mathbb{R}^{M \times M}$ is computed similarly using $\hat{Z}$.
Prediction uncertainty at test points is derived from the posterior covariance matrix
\begin{equation}\label{eq:uncert}
\Sigma = K_{\hat{Z}, \hat{Z}} - K_{\hat{Z}, Z} K^{-1} K_{\hat{Z}, Z}^\top,
\end{equation}
where the diagonal elements represent the prediction variances at the test points.

Hyperparameter optimization is crucial to balance model flexibility and smoothness \cite{Kocijan2015}. 
This is typically achieved by minimizing the negative log-likelihood of the training data
\begin{equation}\label{eq:loglike}
	\log \mathcal{L} \left(l, \nu, \sigma^2 \right) = -\frac{1}{2}\log\left(|K| \right) - \frac{1}{2} \mathbf{y}^\top K^{-1} \mathbf{y} - \frac{N}{2}\log \left(2 \pi \right).
\end{equation}
Minimization is performed by calculating the negative gradient of Eq.~\ref{eq:loglike} with respect to each hyperparameter and iteratively updating them using the Adam optimizer.

\subsection{Tiled algorithms}\label{parallel}
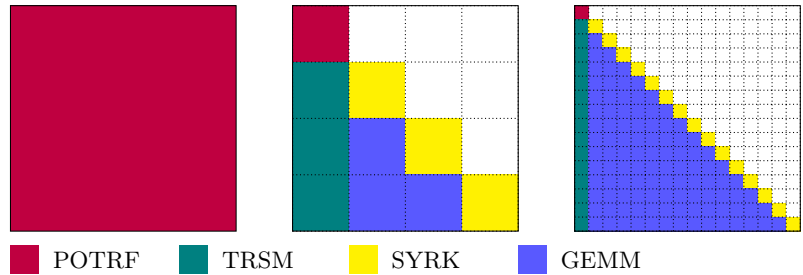
\begin{figure}[t]
    \centering
    \begin{tikzpicture}[scale=0.75] 
        \begin{scope}
            \fill [purple] (0,0) rectangle (4,4);
            \draw (0,0) rectangle (4,4); 
        \end{scope}

        \begin{scope}[xshift=5cm]
            \fill[purple] (0,3) rectangle (1,4);
            \fill[white] (1,3) rectangle (2,4);
            \fill[white] (2,3) rectangle (3,4);
            \fill[white] (3,3) rectangle (4,4);
            
            \fill[teal] (0,2) rectangle (1,3);
            \fill[yellow] (1,2) rectangle (2,3);
            \fill[white] (2,2) rectangle (3,3);
            \fill[white] (3,2) rectangle (4,3);
            
            \fill[teal] (0,1) rectangle (1,2);
            \fill[blue!65] (1,1) rectangle (2,2);
            \fill[yellow] (2,1) rectangle (3,2);
            \fill[white] (3,1) rectangle (4,2);
            
            \fill[teal] (0,0) rectangle (1,1);
            \fill[blue!65] (1,0) rectangle (2,1);
            \fill[blue!65] (2,0) rectangle (3,1);
            \fill[yellow] (3,0) rectangle (4,1);

            \draw (0,0) rectangle (4,4); 
            \draw[step=1cm, densely dotted] (0,0) grid (4,4);
        \end{scope}

        \begin{scope}[xshift=10cm]
            \fill[purple] (0,3.75) rectangle (0.25,4);
            \fill[white] (0.25,3.75) rectangle (4,4);

            \fill[teal] (0,3.5) rectangle (0.25,3.75);
            \fill[yellow] (0.25,3.5) rectangle (0.5,3.75);
            \fill[white] (0.5,3.5) rectangle (4,3.75);

            \fill[teal] (0,3.25) rectangle (0.25,3.5);
            \fill[blue!65] (0.25,3.25) rectangle (0.5,3.5);
            \fill[yellow] (0.5,3.25) rectangle (0.75,3.5);
            \fill[white] (0.75,3.25) rectangle (4,3.5);

            \fill[teal] (0,3) rectangle (0.25,3.25);
            \fill[blue!65] (0.25,3) rectangle (0.5,3.25);
            \fill[blue!65] (0.5,3) rectangle (0.75,3.25);
            \fill[yellow] (0.75,3) rectangle (1,3.25);
            \fill[white] (1,3) rectangle (4,3.25);

            \fill[teal] (0,2.75) rectangle (0.25,3);
            \fill[blue!65] (0.25,2.75) rectangle (0.5,3);
            \fill[blue!65] (0.5,2.75) rectangle (0.75,3);
            \fill[blue!65] (0.75,2.75) rectangle (1,3);
            \fill[yellow] (1,2.75) rectangle (1.25,3);
            \fill[white] (1.25,2.75) rectangle (4,3);

            \fill[teal] (0,2.5) rectangle (0.25,2.75);
            \fill[blue!65] (0.25,2.5) rectangle (0.5,2.75);
            \fill[blue!65] (0.5,2.5) rectangle (1.25,2.75);
            \fill[yellow] (1.25,2.5) rectangle (1.5,2.75);
            \fill[white] (1.5,2.5) rectangle (4,2.75);

            \fill[teal] (0,2.25) rectangle (0.25,2.5);
            \fill[blue!65] (0.25,2.25) rectangle (0.5,2.5);
            \fill[blue!65] (0.5,2.25) rectangle (1.5,2.5);
            \fill[yellow] (1.5,2.25) rectangle (1.75,2.5);
            \fill[white] (1.75,2.25) rectangle (4,2.5);

            \fill[teal] (0,2) rectangle (0.25,2.25);
            \fill[blue!65] (0.25,2) rectangle (0.5,2.25);
            \fill[blue!65] (0.5,2) rectangle (1.75,2.25);
            \fill[yellow] (1.75,2) rectangle (2,2.25);
            \fill[white] (2,2) rectangle (4,2.25);

            \fill[teal] (0,1.75) rectangle (0.25,2);
            \fill[blue!65] (0.25,1.75) rectangle (0.5,2);
            \fill[blue!65] (0.5,1.75) rectangle (2,2);
            \fill[yellow] (2,1.75) rectangle (2.25,2);
            \fill[white] (2.25,1.75) rectangle (4,2);

            \fill[teal] (0,1.5) rectangle (0.25,1.75);
            \fill[blue!65] (0.25,1.5) rectangle (0.5,1.75);
            \fill[blue!65] (0.5,1.5) rectangle (2.25,1.75);
            \fill[yellow] (2.25,1.5) rectangle (2.5,1.75);
            \fill[white] (2.5,1.5) rectangle (4,1.75);

            \fill[teal] (0,1.25) rectangle (0.25,1.5);
            \fill[blue!65] (0.25,1.25) rectangle (0.5,1.5);
            \fill[blue!65] (0.5,1.25) rectangle (2.5,1.5);
            \fill[yellow] (2.5,1.25) rectangle (2.75,1.5);
            \fill[white] (2.75,1.25) rectangle (4,1.5);

            \fill[teal] (0,1) rectangle (0.25,1.25);
            \fill[blue!65] (0.25,1) rectangle (0.5,1.25);
            \fill[blue!65] (0.5,1) rectangle (2.75,1.25);
            \fill[yellow] (2.75,1) rectangle (3,1.25);
            \fill[white] (3,1) rectangle (4,1.25);

            \fill[teal] (0,0.75) rectangle (0.25,1);
            \fill[blue!65] (0.25,0.75) rectangle (0.5,1);
            \fill[blue!65] (0.5,0.75) rectangle (3,1);
            \fill[yellow] (3,0.75) rectangle (3.25,1);
            \fill[white] (3.25,0.75) rectangle (4,1);

            \fill[teal] (0,0.5) rectangle (0.25,0.75);
            \fill[blue!65] (0.25,0.5) rectangle (0.5,0.75);
            \fill[blue!65] (0.5,0.5) rectangle (3.25,0.75);
            \fill[yellow] (3.25,0.5) rectangle (3.5,0.75);
            \fill[white] (3.5,0.5) rectangle (4,0.75);

            \fill[teal] (0,0.25) rectangle (0.25,0.5);
            \fill[blue!65] (0.25,0.25) rectangle (0.5,0.5);
            \fill[blue!65] (0.5,0.25) rectangle (3.5,0.5);
            \fill[yellow] (3.5,0.25) rectangle (3.75,0.5);
            \fill[white] (3.75,0.25) rectangle (4,0.5);

            \fill[teal] (0,0) rectangle (0.25,0.25);
            \fill[blue!65] (0.25,0) rectangle (0.5,0.25);
            \fill[blue!65] (0.5,0) rectangle (3.75,0.25);
            \fill[yellow] (3.75,0) rectangle (4,0.25);

            \draw (0,0) rectangle (4,4); 
            \draw[step=0.25cm, densely dotted] (0,0) grid (4,4); 
        \end{scope}

        \begin{scope}[yshift=-0.75cm]
            \fill [purple] (0,0) rectangle (0.5,0.5);
            \node[anchor=west] at (0.6,0.25) {POTRF};
            
            \fill [teal] (3,0) rectangle (3.5,0.5);
            \node[anchor=west] at (3.6,0.25) {TRSM};
            
            \fill [yellow] (6,0) rectangle (6.5,0.5);
            \node[anchor=west] at (6.6,0.25) {SYRK};
            
            \fill [blue!65] (9,0) rectangle (9.5,0.5);
            \node[anchor=west] at (9.6,0.25) {GEMM};
        \end{scope}
    \end{tikzpicture}
    \caption{Illustration of the first step of a tiled Cholesky decomposition\cite{Buttari2008} using BLAS operations: The left panel shows a single tile, the middle panel shows 4 tiles per dimension, and the right panel shows 16 tiles per dimension, with the BLAS operations POTRF, TRSM, GEMM, and SYRK.}
    \label{fig:tiled_alg}
\end{figure}

In practice, a numerically stable approach to compute the inverse of the symmetric and positive definite covariance matrix $K$ in Eq.~\ref{eq:pred} and Eq.~\ref{eq:uncert} is the Cholesky decomposition. 
This method returns the Cholesky factor, which can be reused to compute both predictions and prediction uncertainties.  
This work is based on the framework introduced in \cite{Strack2023}, which implemented tiled Cholesky decomposition and prediction computation.
The framework utilizes parallel algorithms on a tiled version of the problem. 
Before computations begin, matrices, such as the covariance matrix, are divided into tiles (sub-matrices).
Figure~\ref{fig:tiled_alg} illustrates the first step of a tiled Cholesky decomposition for different matrix decompositions: The left panel shows a single tile, the middle panel depicts 4 tiles per dimension, and the right panel shows 16 tiles per dimension.
Afterward, we execute sequential BLAS operations on matrix tiles instead of the entire matrix, such as Cholesky decomposition (POTRF), triangular matrix solve (TRSM), symmetric rank-k update (SYRK), and general matrix-matrix multiplication (GEMM).
The tiled algorithm avoids unnecessary computations, but leads to an inhomogeneous workload. This is a challenge for efficient parallelization, and the reason why we aim for asynchronous tasks here.
In contrast, libraries like GPyTorch and GPflow rely exclusively on the internal OpenMP parallelization provided by the BLAS library.

\subsection{GPRat}\label{sec:binding}
The GPRat library enables users to create GP objects with attributes such as kernel parameters, trainable parameter flags, training tile configurations, and the number of OS threads. 
It provides functions for predictions with or without uncertainty, hyperparameter optimization, loss computation, and utilities to manage the HPX runtime.

For the first time, the HPX runtime can be started and terminated within a Python environment. This was made possible using pybind11, which creates the bindings for the corresponding \CPP functions.
For initialization, a wrapper function appends the number of OS threads to the input argument vector, converts it into an array of C-style strings, and configures the HPX runtime to initialize with the specified number of OS threads.  
Termination involves two steps: 
First, \texttt{hpx::finalize()} is called to ensure that the runtime stops only after all tasks were executed.
Second, \texttt{hpx::stop()} is called as final HPX function, blocking until the runtime is terminated.
Finally, to ensure correct handling of HPX futurization and asynchronous tasking, \texttt{hpx::threads::run\_as\_hpx\_thread()} is used. 
When a function like \texttt{predict()} is called, \texttt{run\_as\_hpx\_thread()} registers a new task to the HPX runtime.
As the main thread does not automatically become an HPX thread after manual initialization, this ensures execution within the HPX runtime.
By blocking until the function completes, \texttt{run\_as\_hpx\_thread()} maintains synchronization and execution order.

\section{Results}\label{sec:results}

The results section is divided into two parts. 
First, we compare the hyperparameter optimization and prediction performance of GPRat with two reference implementations that use GPflow and GPyTorch, respectively, in a strong scaling experiment.
Second, we examine the problem size scaling for the given application.  
As GPs are widely applied in system identification, we use data from a nonlinear mass-spring-damper simulation. 
In system identification, the feature vectors contain lagged system states, the so-called regressors. 
Depending on the modeling approach and the available data, different numbers of regressors are required to obtain satisfactory results. 
For the runtimes presented in Figures \ref{fig:strong_opt} - \ref{fig:problem_size_pred_full}, we choose a moderate number of eight regressors.
However, we conduct benchmarks for up to 128 regressors.
For our scaling benchmarks, we use an AMD EPYC $7742$
CPU with $64$ physical cores and $256$ MB L3 cache. 
The runtimes of our strong scaling benchmarks are averaged over $5$ runs, with the 95\% percentile confidence error shown. For our problem size scaling benchmarks, the results are averaged over $50$ runs for runtimes smaller than $10^2$ seconds and averaged over $5$ runs for larger runtimes.
All benchmarks use double precision BLAS routines. 
For reference implementations, \texttt{OMP\_PLACES=threads} and \texttt{OMP\_PROC\_BIND=close} were set for optimal thread placement and binding. 
Additional details on the reproducibility of the results are provided in the supplementary material.

\subsection{Strong scaling}\label{sec:strong_scaling}

For our strong scaling benchmarks, the problem size was set to $N$$=$$2^{13}$ training and $M$$=$$2^{13}$ test samples. 
We observe no significant binding overhead for GPRat with Python bindings compared to running GPRat in pure \CPP mode.
This absence of binding overhead is achieved by zero-copy interoperability between \CPP and Python and the lightweight, efficient design of pybind11.

\begin{figure}
    \centering
    \begin{minipage}[t]{.47\textwidth}
        \centering
        \includegraphics[width=\linewidth]{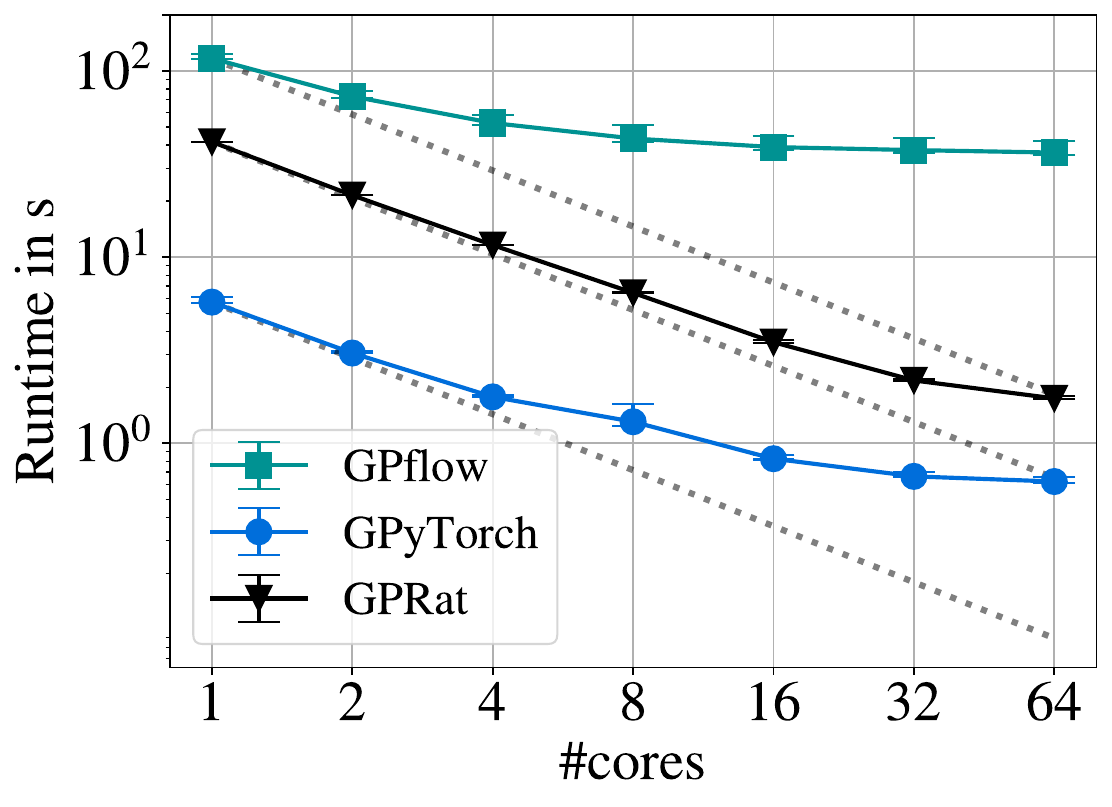}
        \caption{Strong scaling runtimes for hyperparameter optimization on up to $64$ cores. The problem size was set to $N$$=$$2^{13}$ training samples. GPRat uses 16 tiles per dimension (see Figure \ref{fig:tiled_alg}) and yield identical runtimes.} 
        \label{fig:strong_opt}
    \end{minipage}\hspace{.05\textwidth}
    \begin{minipage}[t]{.47\textwidth}
        \centering
        \includegraphics[width=\linewidth]{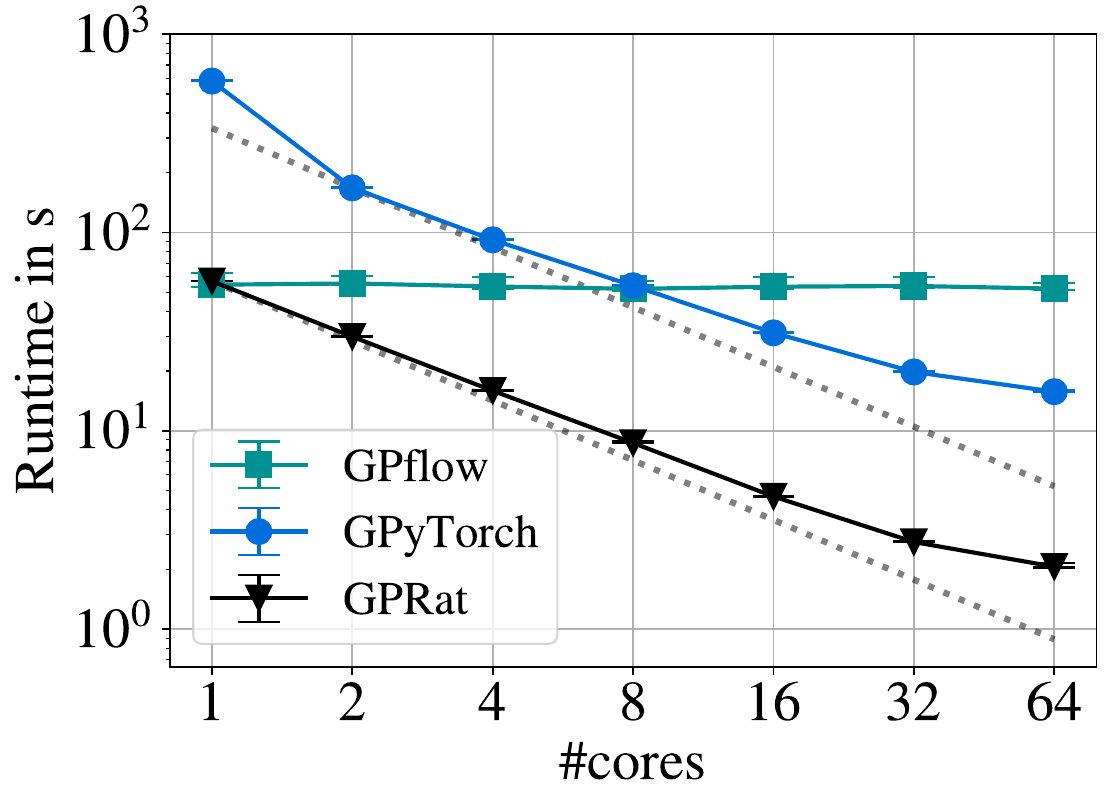}
        \caption{Strong scaling runtimes for prediction with full covariance matrix on up to $64$ cores. The problem size was set to $N$$=$$M$$=$$2^{13}$ training samples and test samples. GPRat uses 16 tiles per dimension (see Figure \ref{fig:tiled_alg}) and yield identical runtimes.} 
        \label{fig:strong_pred}
    \end{minipage}
\end{figure}

Although the GPRat optimization runtime in Figure \ref{fig:strong_opt} is slower than that of GPyTorch, GPRat achieves a parallel speedup of $23.89$ on $64$ cores. Ideal scaling is visualized as a dotted gray line. In comparison, GPyTorch only achieves a parallel speedup of $9.21$.
The worse performance of GPRat is related to closed-form gradient calculations. In contrast, GPyTorch leverages PyTorch's backpropagation for gradient accumulation.
Interestingly, there is a noticeable difference between the optimization performance of GPflow and GPyTorch, even though both libraries employ backpropagation. The exact cause of GPflow's inferior performance remains unclear, as it relies on TensorFlow's internal computation graph for backpropagation.  
In general, performance scaling on up to $64$ cores is constrained due to implicit communication overhead, caused by the chiplet design of the AMD EPYC $7742$ CPU \cite{Lane2022}.

Figure \ref{fig:strong_pred} shows the strong scaling benchmark for prediction with full covariance uncertainty. 
GPRat demonstrates similar scaling compared to Figure \ref{fig:strong_opt}, achieving a speedup of $27.54$ for $64$ cores. 
In contrast, GPflow scaling is less efficient because some of the employed functions, such as \texttt{tf.linalg.cholesky()} and \texttt{tf.linalg.triangular\_solve()} are not parallelized on CPUs.
GPyTorch scales better than GPflow but has higher runtimes for the given problem setting. This is attributed to GPyTorch's computation of prediction uncertainty and predictions\footnote{\url{https://github.com/cornellius-gp/gpytorch/blob/develop/gpytorch/models/exact_prediction_strategies.py} (visited on 01/28/2025)}, which do not reuse results from forward substitution. Instead, they introduce an additional backward substitution step that scales with $O(MN^2)$.
GPRat, on the other hand, achieves speedups of $7.63$ over GPyTorch and $25.25$ over GPflow for $64$ cores. 

\begin{figure}
    \centering
    \begin{minipage}[t]{.44\textwidth}
        \centering
        \includegraphics[width=\linewidth]{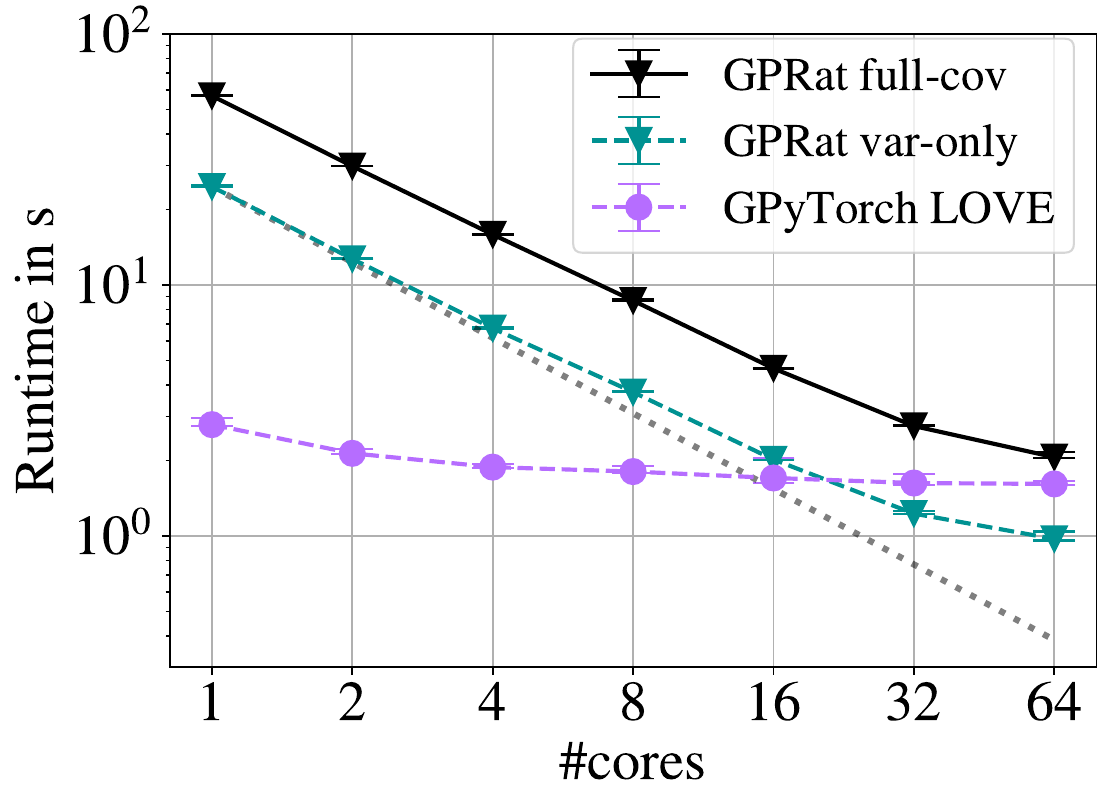}
        \caption{Strong scaling runtimes for prediction with full covariance matrix versus variance only on up to $64$ cores for GPRat and GPyTorch with LOVE.         The problem size was set to $N$$=$$M$$=$$2^{13}$ training samples and test samples. GPRat uses 16 tiles per dimension (see Figure \ref{fig:tiled_alg}).} 
        \label{fig:strong_LOVE}
    \end{minipage}\hspace{.05\textwidth}
    \begin{minipage}[t]{.5\textwidth}
        \centering
        \includegraphics[width=\linewidth]{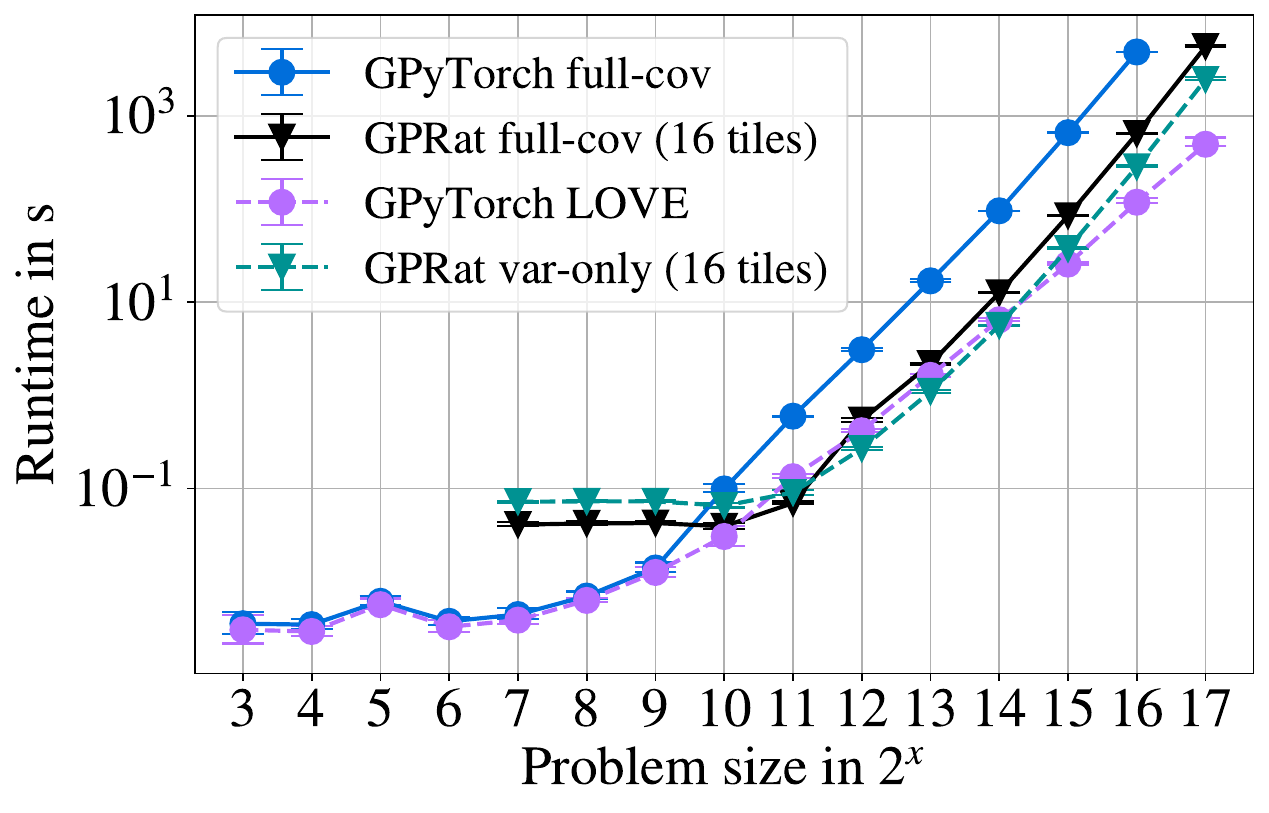}
        \caption{Problem size scaling runtimes for prediction with full covariance matrix versus variance only on $64$ cores for GPRat, GPyTorch, and GPyTorch with LOVE. GPRat uses 16 tiles per dimension (see Figure \ref{fig:tiled_alg}).
        } 
        \label{fig:problem_size_pred_LOVE}
    \end{minipage}
\end{figure}

Lastly, Figure \ref{fig:strong_LOVE} shows the strong scaling benchmark for prediction with full covariance matrix versus variance only, comparing GPRat and GPyTorch with LOVE. 
The LOVE algorithm approximates the predictive covariance matrix and has better asymptotic complexity compared to the naive approach \cite{Pleiss2018}.
GPRat full-cov and var-only show similar scaling. 
However, the var-only variant shows significant performance improvements. 
In comparison, GPyTorch with LOVE has a shorter runtime on fewer than 16 cores. 
Nonetheless, GPRat var-only outperforms the approximation algorithm as the number of cores increases, achieving a speedup of $1.66$ over GPyTorch with LOVE. 
The limited scaling of GPyTorch with LOVE can be related to sequential precomputation steps in the approximation, despite the use of parallelized functions such as \texttt{torch.matmul()}.

\subsection{Problem size scaling}\label{sec:problem_size}

In this subsection, we present the results of our problem size scaling experiments for optimization and prediction with full covariance. 
All runs were performed on 64 cores. 
The problem size, with the same number of training and test samples, is varied from $2^3$ to $2^{17}$ in successive powers of 2 ($2^3,2^4,\ldots,2^{17}$).
For GPRat, results are reported for $4$ and $16$ tiles per dimension, with the one tile, i.e., sequential computation serving as a baseline (compare Figure \ref{fig:tiled_alg}). 

\begin{figure}
    \centering
    \begin{minipage}[t]{.47\textwidth}
        \centering
        \includegraphics[width=\linewidth]{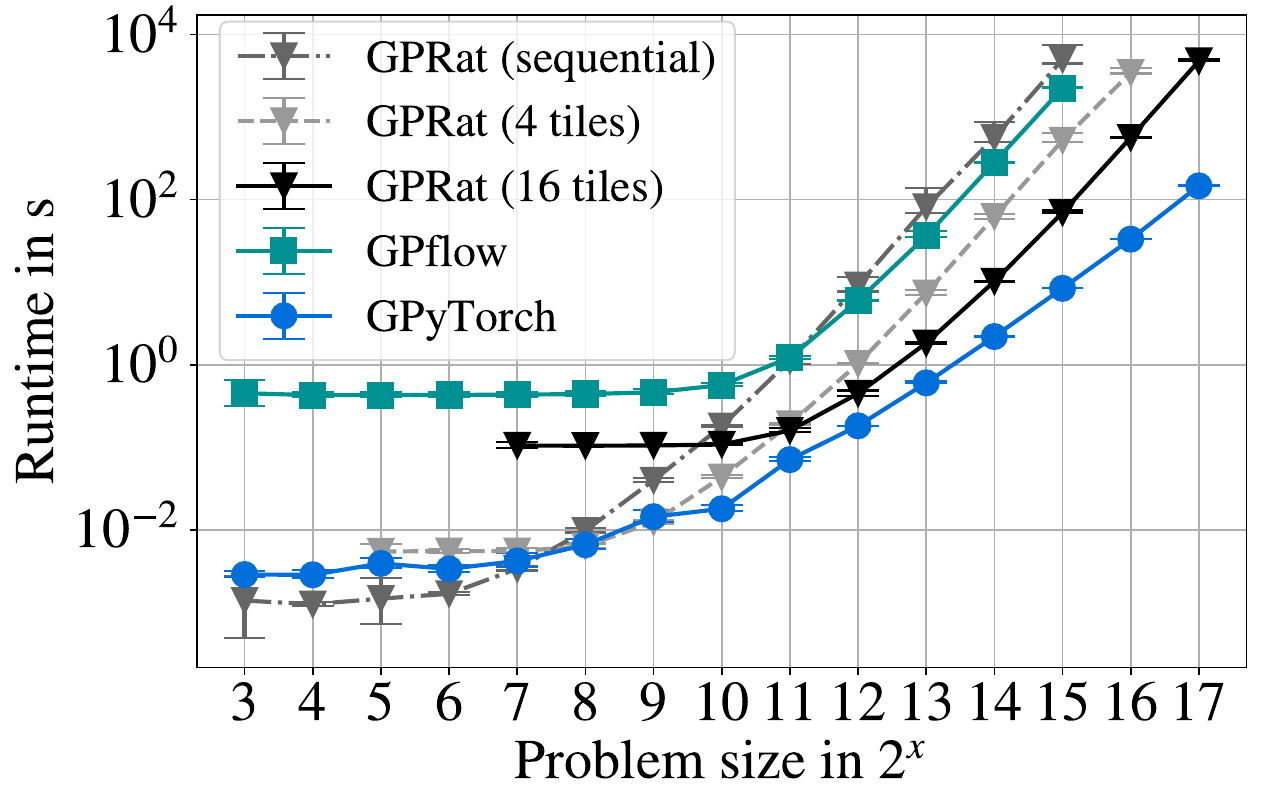}
        \caption{Problem size scaling runtimes for hyperparameter optimization on $64$ cores. 
        For GPRat, the tile size per dimension was set to $1$, $4$, and $16$ (see Figure \ref{fig:tiled_alg}).
        }
        \label{fig:problem_size_opt}
    \end{minipage}\hspace{.05\textwidth}
    \begin{minipage}[t]{.47\textwidth}
        \centering
        \includegraphics[width=\linewidth]{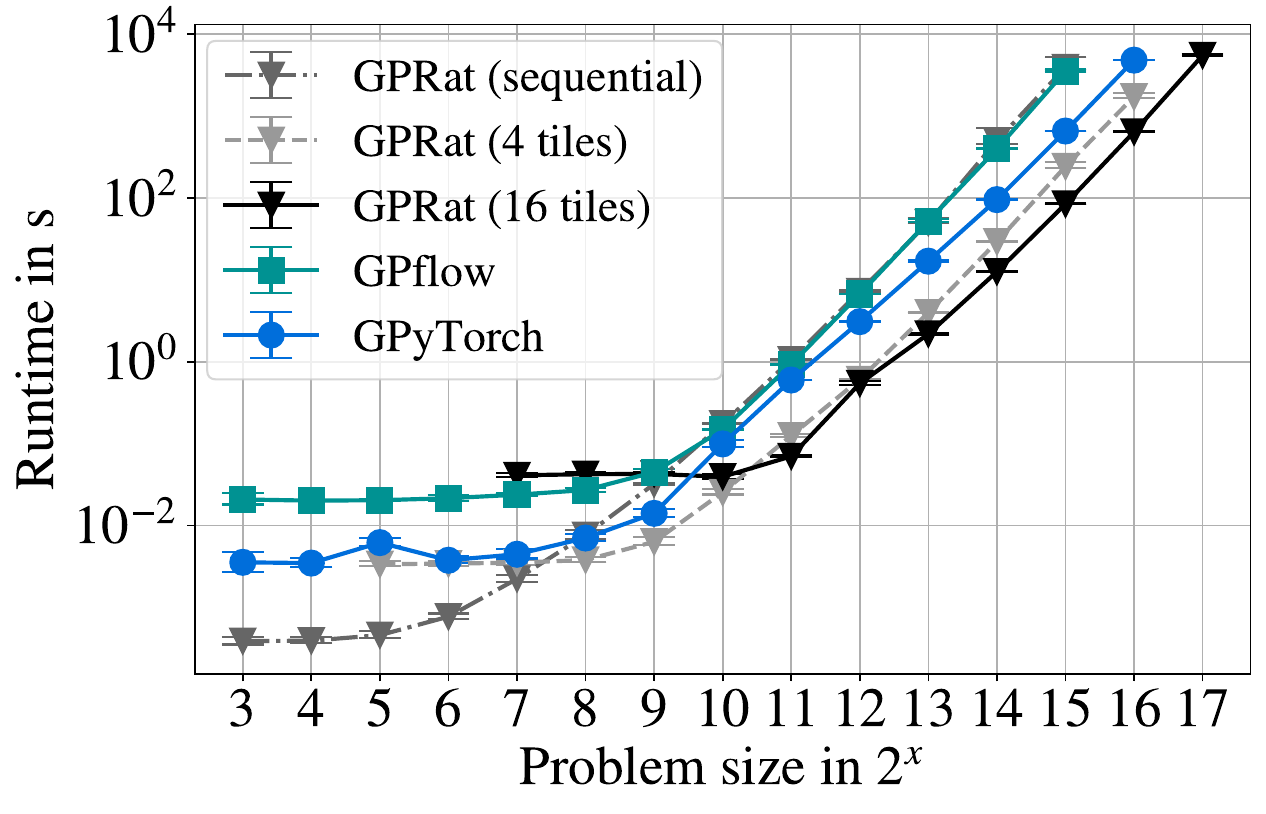}
        \caption{Problem size scaling runtimes for prediction with full covariance matrix on $64$ cores.
        For GPRat, the tile size per dimension was set to $1$, $4$, and $16$ (see Figure \ref{fig:tiled_alg}).
        } 
        \label{fig:problem_size_pred_full}
    \end{minipage}
\end{figure}

Figure \ref{fig:problem_size_opt} shows runtime benchmarks for hyperparameter optimization across multiple problem sizes. 
Runtime increases with problem size for all implementations. 
GPflow shows the largest runtime overhead only exceeding the sequential baseline at $N$$=$$2^{11}$.
The GPRat overhead increases for more tiles per dimension.
For large problem sizes, GPRat with $16$ tiles per dimension is faster than the $4$-tile variant. We can not observe notable performance gains for more tiles per dimension.
GPyTorch yields the shortest runtimes beyond $N$$=$$2^9$ and the best scaling. These results are consistent with the strong scaling benchmark, where GPyTorch excels due to efficient backpropagation.

The problem size scaling runtimes for a prediction with full covariance matrix on $64$ cores for GPRat, GPflow, and GPyTorch are presented in Figure \ref{fig:problem_size_pred_full}. 
Again, the runtime increases with problem size for all implementations.
GPflow shows the largest runtime overhead and is only marginally faster than the sequential version.
GPyTorch only achieves good performance for problem sizes below $N$$=$$M$$=$$2^{10}$.
In contrast, GPRat demonstrates strong scalability with $16$ tiles per dimension, achieving the shortest runtime for problem sizes greater than $N$$=$$M$$=$$2^{10}$. 
This configuration achieves the optimal balance between task size and scheduling effort for both optimization and prediction with full covariance matrix.

Lastly, Figure \ref{fig:problem_size_pred_LOVE} compares problem size scaling runtimes for prediction using full covariance (full-cov) and variance only (var-only) computations on $64$ cores for GPRat and GPyTorch, including GPyTorch with the LOVE approximation. 
GPyTorch with full-cov has the longest runtime, especially beyond $N$$=$$M$$=$$2^{10}$. 
The LOVE approximation significantly reduces the runtime of GPyTorch. Additional runtime reductions are possible by lowering the number of Lanczos iterations (we use the default value). However, this reduction is a trade-off for approximation accuracy.
For GPRat with $16$ tiles per dimension, full-cov and var-only runtimes perform similarly for small problem sizes.
For problem sizes larger than $N$$=$$M$$=$$2^{12}$ the var-only configuration consistently outperforms its full-cov counterpart. In our benchmark, GPRat with var-only was even able to outperform the LOVE approximation until the inferior algorithmic complexity shows its toll.

Note that the performance of GPyTorch is highly sensitive to the number of regressors. For 128 regressors, GPyTorch's prediction with full-cov shows only minor performance improvements compared to the sequential version. In addition, the LOVE approximation is slower than GPRat with var-only across the board. This indicates an inefficient assembly of the covariance matrix. 

\section{Conclusion and outlook}\label{sec:conclusion}

In this work, we introduce GPRat, a novel task-based GP library for \CPP and Python. In addition, we propose a new method to integrate the HPX runtime into Python using pybind11.
Furthermore, we conduct node-level scaling benchmarks comparing GPRat against two reference implementations based on GPflow and GPyTorch on up to 64 cores.
While GPRat leverages asynchronous tasking, GPflow and GPyTorch rely on OpenMP for CPU parallelization.

The benchmarks indicate that there is no significant binding overhead when using HPX within a Python API using pybind11.
For our benchmark hardware configuration, the optimal number of tiles per dimension is $16$. 
GPRat's hyperparameter optimization performance cannot keep up with GPyTorch's efficient backpropagation. 
However, GPRat shows better scaling. 
Furthermore, GPRat delivers significant prediction advantages, achieving up to $7.63\times$ faster predictions on $64$ cores.
Our analysis reveals that GPflow fails to scale effectively on CPUs due to the lack of parallelization in key TensorFlow operations, like Cholesky decomposition and triangular solve. 
In addition, we demonstrate that GPRat can outperform GPyTorch with LOVE by a factor of up to $1.66$, despite the latter using an algorithm with superior asymptotic complexity.
This highlights the potential of GPRat for substantial performance gains in online prediction, particularly in environments with high computational resources.

In future work, we plan to implement a backpropagation algorithm, similar to PyTorch, to explore its scalability within our task-based asynchronous framework. 
Moreover, our goal is to enable GPU support using HPX accelerator executors, as well as to extend GPRat for distributed computing across multiple nodes and clusters.  
We also plan to evaluate the performance portability of GPRat, GPflow, and GPyTorch on different hardware architectures, specifically ARM and RISC-V.
Finally, additional features may include more kernel functions (e.g., Mat\'{e}rn kernels), optimization algorithms, support for sparse GPs, and automatic tile size selection.

\section*{Supplementary materials}
The GPRat library is open source under the MIT license and is available on \href{https://github.com/SC-SGS/GPRat}{GitHub}\footnote{\url{https://github.com/SC-SGS/GPRat} (visited on 03/12/2025)}. 
Additional information and material to reproduce the results presented in this work are available on \href{https://doi.org/10.18419/DARUS-4743}{DaRUS}\footnote{\url{
https://doi.org/10.18419/DARUS-4743} (visited on 03/12/2025)}.

\bibliographystyle{splncs04}
\bibliography{main}
\end{document}